\documentclass[runningheads]{llncs}
\usepackage[T1]{fontenc}
\usepackage{graphicx}
\usepackage{amsfonts}
\usepackage{siunitx}
\usepackage{multirow}
\usepackage{hyperref}
\usepackage{url}

\usepackage[table]{xcolor} 
\usepackage{booktabs}      
\newcommand{\first}[1]{\cellcolor[HTML]{63C07A}#1}   
\newcommand{\second}[1]{\cellcolor[HTML]{BDDC8E}#1}  
\newcommand{\third}[1]{\cellcolor[HTML]{EBE7A0}#1}            

\usepackage[most]{tcolorbox} 
\usepackage[dvipsnames, table]{xcolor}

\definecolor{firstColor}{HTML}{63C07A}
\definecolor{secondColor}{HTML}{BDDC8E}
\definecolor{thirdColor}{HTML}{EBE7A0}

\newcommand{\First}[1]{\colorbox{firstColor}{#1}}
\newcommand{\Second}[1]{\colorbox{secondColor}{#1}}
\newcommand{\Third}[1]{\colorbox{thirdColor}{#1}}

\begin{document}


\title{MambaKick: Early Penalty Direction Prediction from HAR Embeddings}



%
%

\author{Henry O. Velesaca\inst{1,2} \and
David Freire-Obreg\'on\inst{3} \and
Abel Reyes-Angulo\inst{4} \and \\
Steven Araujo\inst{1} \and 
Angel Sappa\inst{1,5}}

\authorrunning{Velesaca et al.}
%

\institute{ESPOL Polytechnic University, ESPOL, Campus Gustavo Galindo, Km. 30.5 Vía Perimetral, Guayaquil, 090902, Ecuador \\ \email{\{hvelesac,saraujo,asappa\}@espol.edu.ec} \and
Software Engineering Department, Research Center for Information and Communication Technologies (CITIC-UGR), University of Granada, 18071, Granada, Spain
\email{hvelesaca@correo.ugr.es} \and
SIANI, Universidad de Las Palmas de Gran Canaria, Spain \email{david.freire@ulpgc.es} \and
Michigan Technological University, 49931, Houghton, MI, EEUU \\ \email{areyesan@mtu.edu} \and
Computer Vision Center, 08193-Bellaterra, Barcelona, Spain\\
\email{sappa@ieee.org}}

\maketitle              
\begin{abstract}
Penalty kicks in soccer are decided under extreme time constraints, where goalkeepers benefit from anticipating shot direction from the kicker’s motion before or around ball contact. In this paper, \textit{MambaKick} is presented as a learning-based framework for penalty direction prediction that leverages pretrained human action recognition (HAR) embeddings extracted from contact-centered $short $video segments and combines them with a lightweight temporal predictor. Rather than relying on explicit kinematic reconstruction or handcrafted biomechanical features, the approach reuses transferable spatiotemporal representations and utilizes selective state-spare models (Mamba) for efficient sequence aggregation. Simple contextual metadata (e.g., field side and footedness) are also considered as complementary cues that may reduce ambiguity in real-world footage. Across a range of HAR backbones, \textit{MambaKick} consistently improves or matches strong embedding baselines, achieving up to 53.1\% accuracy for three classes and 64.5\% for two classes under the proposed methodology. Overall, the results indicate that combining pretrained HAR representations with efficient state-space temporal modeling is a practical direction for low-latency intention prediction in real-world sports video. The code will be available at GitHub: \url{https://github.com/hvelesaca/MambaKick/}.


\end{abstract}
\section{Introduction}
Penalty kicks are among the most decisive and time-constrained events in soccer: the ball is struck from 11 meters, leaving goalkeepers only a fraction of a second to react. Consequently, a key factor for successful goalkeeping is \emph{anticipation}—initiating movement before (or very close to) ball contact—rather than purely reactive diving. Empirical analyses and controlled studies have shown that goalkeeping behavior in penalties is shaped by a combination of time constraints, strategic biases, and perceptual cues extracted from the kicker’s run-up and kinematics \cite{BarEli2007ActionBias,VanderKamp2006LateAlterations}. In particular, the ability to infer the intended shot direction early, from pre-contact body configuration, can increase the effective time available to intercept the ball.

Prior work in sports biomechanics and perceptual-motor research has identified informative pre-contact cues for kick direction, such as support-foot orientation and approach angle \cite{LeesOwens2011EarlyVisualCues,Li2015PredictionKickDirection}. However, translating these findings into an automated system that operates robustly on real video remains challenging. Broadcast conditions introduce viewpoint changes, occlusions, and large appearance variability, and many classical pipelines depend on accurate pose/kinematic reconstruction or handcrafted measurements—components that are often brittle outside laboratory settings.

In parallel, computer vision has made substantial progress in \emph{human action recognition} (HAR) by learning transferable video representations from large-scale data. Modern backbones based on 3D convolutional networks and factorized spatiotemporal models learn embeddings that capture motion and posture patterns \cite{CarreiraZisserman2017QuoVadis,Feichtenhofer2019SlowFast}, while video Transformers extend this capability through attention-based spatiotemporal modeling \cite{Bertasius2021TimeSformer}. These pretrained HAR models provide a practical way to convert short video clips into compact feature vectors that can serve as \emph{embeddings} for downstream prediction tasks, including fine-grained sports actions.

This paper introduces \textit{MambaKick}, a framework for penalty direction prediction from HAR embeddings extracted from a contact-centered temporal window. The pre-contact segment captures anticipatory cues needed for early intention inference, while the short post-contact tail helps stabilize the representation around impact and reduces sensitivity to small alignment errors in contact annotation. In addition, we incorporate lightweight contextual metadata—such as pitch side and the kicker’s dominant foot—which can affect visual appearance and directional priors in broadcast settings, and can therefore complement the visual signal.

Our contributions are threefold: (i) we formulate penalty direction prediction using a compact contact-centered window of HAR embeddings complemented by contextual metadata; (ii) we propose \textit{MambaKick}, combining pretrained HAR representations with a Mamba-based \cite{GuDao2023Mamba} temporal head for efficient sequence modeling; and (iii) we evaluate the approach in both three-class (left/center/right) and two-class (left/right) settings, reporting accuracy and class-wise behavior.

The manuscript is organized as follows. Section~\ref{sec:relwork} provides a review of related work. Sections~\ref{sec:method} and~\ref{sec:expSet} detail the proposed methodology and the corresponding experimental setup, respectively. Section~\ref{sec:expRes} summarizes the experimental results, comparative analyzes, and discussion. Finally, Section~\ref{sec:concl} summarizes the main conclusions of the study.

\section{Related Work}
\label{sec:relwork}
Penalty kicks have been widely studied as a dyadic interaction under severe time pressure, where successful goalkeeping depends more on anticipation than on purely reactive movement. Analyses of elite competitions highlight systematic behavioral tendencies, such as action bias and the limitations of late reactions \cite{BarEli2007ActionBias}. Controlled experiments show that late alterations in kick direction increase execution errors and reduce accuracy, implying a narrow temporal window in which intention becomes stable \cite{VanderKamp2006LateAlterations}. From a biomechanical perspective, prior work identifies informative pre-contact cues (e.g., support-foot orientation and approach variables) that correlate with final direction \cite{LeesOwens2011EarlyVisualCues,Li2015PredictionKickDirection}. Although these studies provide interpretable factors, their direct operationalization in real broadcast footage remains difficult: cue extraction typically requires accurate measurements, consistent viewpoints, and reliable tracking, which are often violated by occlusions, camera motion, and appearance variability in the wild.

In parallel, soccer-focused benchmarks have enabled learning-based video understanding at scale. SoccerNet provides large-scale annotations for event spotting in broadcast videos \cite{Giancola2018SoccerNet}, and subsequent work proposes context-aware objectives to improve spotting of sparse events in long temporal streams \cite{Cioppa2020CALF}. These datasets and methods contribute valuable supervision and temporal reasoning tools; however, the primary objective is commonly long-form event localization rather than fine-grained, pre-contact intention inference. As a result, they do not directly address the early-prediction constraint that arises in penalty kicks, where decisions must be made from partial observations under strict latency.

A practical alternative to handcrafted biomechanical pipelines is to reuse pretrained human action recognition (HAR) models as feature extractors, leveraging representations learned from large-scale video data. Architectures such as I3D \cite{CarreiraZisserman2017QuoVadis} and SlowFast \cite{Feichtenhofer2019SlowFast} encode spatiotemporal patterns that transfer effectively to downstream tasks, while video Transformers further increase representational capacity through space--time attention \cite{Bertasius2021TimeSformer}. Despite these advantages, HAR features are often consumed via clip-level pooling and shallow classifiers, which can obscure how intent evolves across the run-up and may be sensitive to temporal alignment around contact. In addition, high-capacity attention-based models can introduce non-trivial computational overhead, which is undesirable in low-latency decision settings.

Recent studies have explored human-centric video representations in soccer analytics. Gait-based embeddings have been used to classify penalty shooting zones~\cite{Freireics25}, showing informative cues but relying on full sequences and identity-oriented models, which limits their value for action prediction. HAR embeddings have also been examined in free-kick scenarios, demonstrating that pre-kick motion can predict ball direction~\cite{Toron23}, although three-class predictions introduce ambiguity in the center-class outcome. Later work extends this idea via heterogeneous transfer learning for gender and shot direction prediction in free kicks~\cite{Toron24}. While promising, these efforts target free kicks rather than penalties and use coarse temporal segmentation, leaving open how motion patterns evolve during the penalty run-up.

Temporal modeling choices further shape early-prediction performance. Recurrent models such as LSTMs provide a classical approach to sequence learning \cite{Hochreiter1997LSTM}, while Transformers offer strong accuracy through attention-based aggregation \cite{Vaswani2017Attention}. Nonetheless, both paradigms can be suboptimal for streaming inference: recurrent models may struggle to preserve fine-grained discriminative cues over longer contexts, and Transformers typically incur higher computational cost as sequence length grows. Selective state-space models, as instantiated by Mamba, address these limitations by enabling linear-time sequence modeling with competitive performance \cite{GuDao2023Mamba}, making them attractive for efficient, online temporal reasoning over frame-level or embedding-level sequences \cite{velesaca2025samba}.

Pose-based formulations provide another direction by explicitly representing body configuration. Systems such as OpenPose enable keypoint extraction in unconstrained scenes \cite{Cao2017OpenPose}, and skeleton-based recognition methods such as ST-GCN operate directly on joint trajectories \cite{Yan2018STGCN}. These approaches can improve interpretability and reduce sensitivity to appearance, but they remain vulnerable to detection noise, occlusions, and viewpoint changes---conditions that are common in penalty footage and can propagate errors through the pipeline. Moreover, reliable multi-person pose estimation and tracking can increase complexity and reduce robustness in real deployments.

Within this landscape, a baseline for penalty-kick direction prediction has recently been proposed, leveraging pretrained HAR backbones and their embeddings for classification \cite{freire2025predicting}. Building upon this embedding-based foundation, \textit{MambaKick} is designed to better exploit the temporal structure of HAR embeddings via an efficient Mamba-based head and to optionally incorporate lightweight contextual metadata, thereby improving robustness and practicality for low-latency intention prediction in real-world sports video \cite{GuDao2023Mamba}.

\begin{figure}
    \centering
    \includegraphics[width=\textwidth]{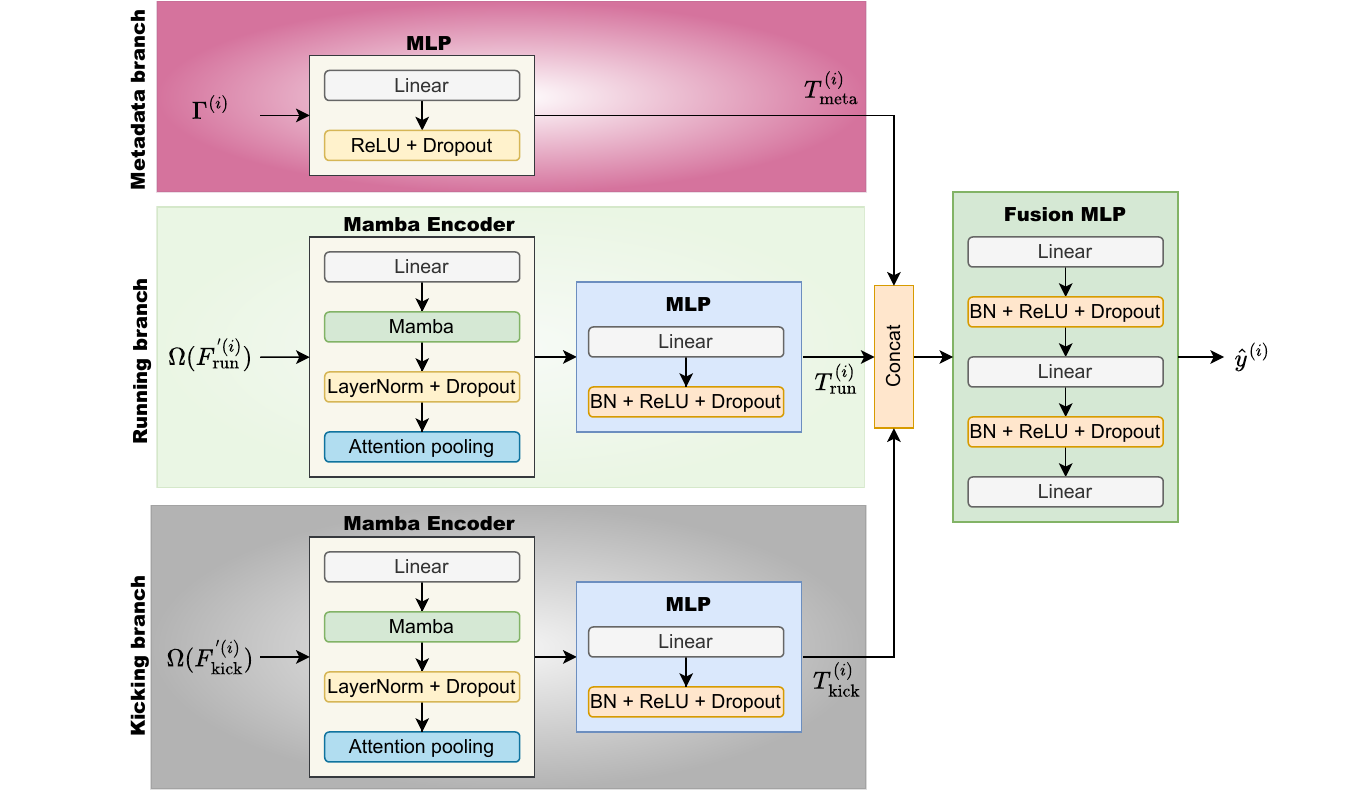}
    \caption{Overview of the proposed MambaKick architecture. The model consists of three branches processing metadata, running-phase dynamics, and kicking-phase dynamics, where visual features are temporally encoded using Mamba encoders and attention pooling before late fusion for shot direction prediction.}
    \label{fig:mambakick}
\end{figure}

\section{Methodology}
\label{sec:method}

Let $\mathcal{S}=\{(F^{(i)}, y^{(i)},\Gamma^{(i)}\}_{i=1}^{m}$ be a dataset that comprises $m$ penalty-kick samples. Each instance consists of a video recording $F^{(i)}$ of the full kicking action, a category label $y^{(i)}$ specifying the final shot direction, and a set of auxiliary attributes $\Gamma^{(i)}$ available before the execution. In this work, $\Gamma^{(i)}$ includes two binary descriptors: the pitch side from which the penalty is taken, and the dominant kicker's foot.

In this work, two prediction tasks are considered: a three-class setup with $y^{(i)} \in \{\textit{left}, \textit{center}, \textit{right}\}$, and a reduced binary version where $y^{(i)} \in \{\textit{left}, \textit{right}\}$. The aim is to learn the parameters $\theta$ of a multi-branch neural network that leverages visual motion cues along with metadata to infer the shot direction. Consequently, training is performed by minimizing the empirical cross-entropy loss:
\begin{equation}
\mathcal{L}(\theta) = -\frac{1}{m} \sum_{i=1}^{m}\sum_{k=1}^{n} \mathbb{I} (y^{(i)} = k)\log\big(p_\theta(\hat{y}^{(i)}=k)\big)
\end{equation}
where $n$ denotes the number of classes and $p_\theta(\hat{y}^{(i)}=k)$ is the predicted probability for class $k$.

\subsection{Preliminaries: Selective State Space Models}
\label{sec:mamba_theory}
The core temporal modeling in our framework relies on Mamba, a selective state-space model (SSM) architecture \cite{GuDao2023Mamba}. SSMs map a 1D input sequence $x(t) \in \mathbb{R}$ to an output $y(t) \in \mathbb{R}$ through a latent state $h(t) \in \mathbb{R}^N$. This process is modeled by a linear ordinary differential equation (ODE):
\begin{equation}
    h'(t) = \mathbf{A}h(t) + \mathbf{B}x(t), \quad y(t) = \mathbf{C}h(t)
\end{equation}
where $\mathbf{A} \in \mathbb{R}^{N \times N}$ is the evolution parameter and $\mathbf{B} \in \mathbb{R}^{N \times 1}, \mathbf{C} \in \mathbb{R}^{1 \times N}$ are projection parameters.

To operate on discrete video embeddings, the continuous parameters $(\mathbf{A}, \mathbf{B})$ are transformed into discrete parameters $(\bar{\mathbf{A}}, \bar{\mathbf{B}})$ using a timescale parameter $\Delta$. Using the zero-order hold (ZOH) method, the discretized recurrence becomes:
\begin{equation}
    h_t = \bar{\mathbf{A}} h_{t-1} + \bar{\mathbf{B}} x_t, \quad y_t = \mathbf{C} h_t
\end{equation}
where $\bar{\mathbf{A}} = \exp(\Delta \mathbf{A})$ and $\bar{\mathbf{B}} = (\Delta \mathbf{A})^{-1}(\exp(\Delta \mathbf{A}) - \mathbf{I}) \cdot \Delta \mathbf{B}$.

While standard SSMs are time-invariant (parameters are fixed across the sequence), Mamba introduces a \textit{selection mechanism} allowing the model to selectively propagate or forget information based on the current input. Specifically, $\bar{\mathbf{B}}$, $\mathbf{C}$, and $\Delta$ become functions of the input $x_t$:
\begin{equation}
    \bar{\mathbf{B}}_t, \mathbf{C}_t, \Delta_t = \text{Linear}(x_t)
\end{equation}
This data-dependent discretization allows Mamba to compress context effectively, filtering irrelevant noise while maintaining long-range dependencies essential for capturing fine-grained motion cues in the penalty run-up.

\subsection{Multi-Branch Temporal Representation}

The proposed \textit{MambaKick} architecture decomposes the penalty kick into three complementary informative streams. As illustrated in Figure \ref{fig:mambakick}, these branches handle (i) contextual metadata, (ii) visual dynamics of the running phase, and (iii) visual dynamics of the kicking phase. This design allows the model to learn specialized representations for each branch while supporting joint reasoning at a later fusion stage.

After context-constrained pre-processing (see Section \ref{sec:pipeline}), each video $F^{(i)}$ is temporally segmented into two disjoint subsequences:

\[
F'^{(i)} = F_{\text{run}}^{'(i)} \cup F_{\text{kick}}^{'(i)}
\]

Where $F_{\text{run}}^{'(i)}$ captures the approach motion and $F_{\text{kick}}^{'(i)}$ corresponds to the final kicking action. Each is downsampled and split into overlapping clips, which are independently encoded by a frozen human action recognition (HAR) backbone. This yields two ordered sets of clip-level embeddings:

\[
\Omega(F_{\text{run}}^{'(i)}) = \{f_{\text{run}}^{(i,1)}, \dots, f_{\text{run}}^{(i,N_r)}\}, \quad
\Omega(F_{\text{kick}}^{'(i)}) = \{f_{\text{kick}}^{(i,1)}, \dots, f_{\text{kick}}^{(i,N_k)}\}
\]

Each sequence of embeddings is then processed by a Mamba encoder, enabling efficient modeling of long-range temporal dependencies within each phase, running and kicking. Attention-based temporal pooling is applied at the output of each encoder. By doing this, we aim to obtain more compact representations:



\begin{equation}
\begin{aligned}
T_{\text{run}}^{(i)} &= \text{AttnPool}\big(\text{Mamba}(\Omega(F_{\text{run}}^{'(i)}))\big) \\
T_{\text{kick}}^{(i)} &= \text{AttnPool}\big(\text{Mamba}(\Omega(F_{\text{kick}}^{'(i)}))\big)
\end{aligned}
\end{equation}

Specifically, the attention pooling layer learns a weight vector $w \in \mathbb{R}^D$ to compute a score $e_t = w^\top h_t$ for each time step, producing a weighted sum $T_{\text{visual}}^{(i)} = \sum \alpha_t h_t$ where $\alpha_t$ are the softmax-normalized scores.

At the same time, a lightweight MLP processes the metadata vector $\Gamma^{(i)}$ to obtain a latent embedding $T_{\text{meta}}^{(i)}$.

Finally, the three representations are concatenated and passed to a fusion head that produces the final outcome:

\begin{equation}
\hat{y}^{(i)} = FusionMLP\big(Concat\big(T_{\text{run}}^{(i)}, T_{\text{kick}}^{(i)}, T_{\text{meta}}^{(i)}\big)\big)
\end{equation}

The fusion head consists of linear layers followed by Batch Normalization, ReLU activation, and Dropout, ensuring robust non-linear combination of the multimodal features. The overall architecture is depicted in Figure \ref{fig:mambakick}, which highlights the branches processing and the late fusion strategy enabled by the Mamba-based encoders.

\subsection{Pipeline Description}
\label{sec:pipeline}

The proposed pipeline comprises three stages: video pre-processing, temporal encoding, and multi-modal classification.

\begin{figure}
    \centering
    \includegraphics[width=\textwidth]{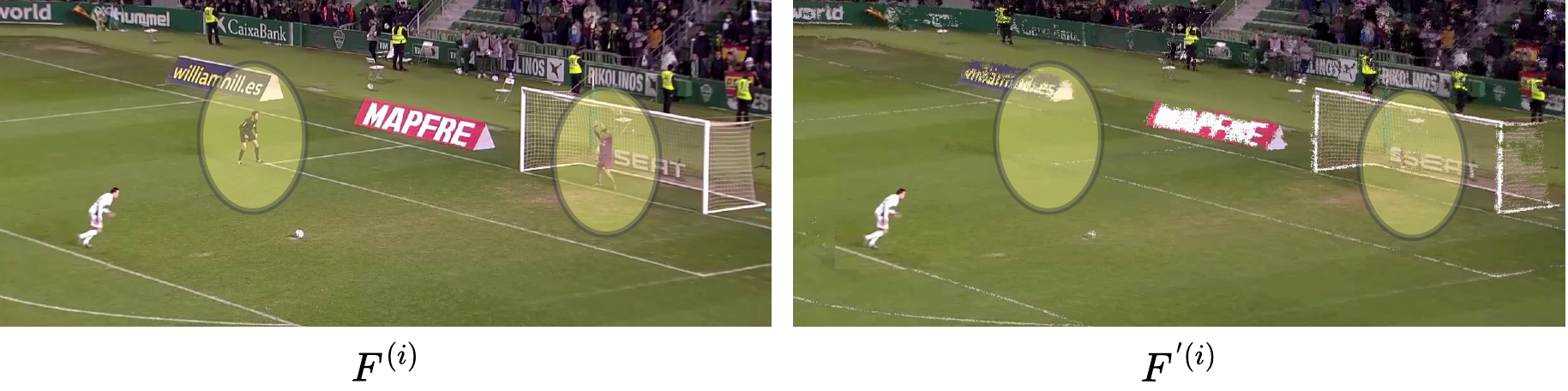}
    \caption{Original frame $F^{(i)}$ (left) and context-constrained frame $F'^{(i)}$ (right), where only the kicker remains as the moving element after background suppression.}
    \label{fig:context_preprocess}
\end{figure}

\textbf{Stage 1: Context-Constrained Pre-Processing}. To suppress irrelevant visual information (i.e., the referee, the crowd, etc.), the kicker is isolated using ByteTrack \cite{zhang2021bytetrack}. For each frame, the kicker's bounding box is overlaid onto a static background resulting from averaging all frames in the sequence. As shown in Figure~\ref{fig:context_preprocess}, this generates a new video where the kicker is the only moving element, allowing a better extraction of the fine-grained motion cues. Moreover, videos are temporally cropped and padded to ensure fixed-length running and kicking segments.

\textbf{Stage 2: Temporal Encoding}. The running and the kicking sequences are independently decomposed into overlapping clips and passed through a pre-trained HAR backbone. Instead of pooling temporal information early, clip embeddings are fed into separate Mamba encoders for each phase, followed by attention pooling. Contrary to prior works, this allows the model to better capture distinct temporal patterns associated with both running dynamics and ball striking mechanics.

\textbf{Multi-Modal Classification}. Metadata features are processed by a dedicated branch, while running and kicking embeddings are refined by their respective phase-specific branches. The resulting latent features are concatenated and passed through a fusion network. This fusion network is composed of fully connected layers with normalization and dropout. Finally, the output layer produces either a three-class or binary prediction of the shot direction, depending on the experimental setting.

\begin{figure}
    \centering
    \includegraphics[width=\textwidth]{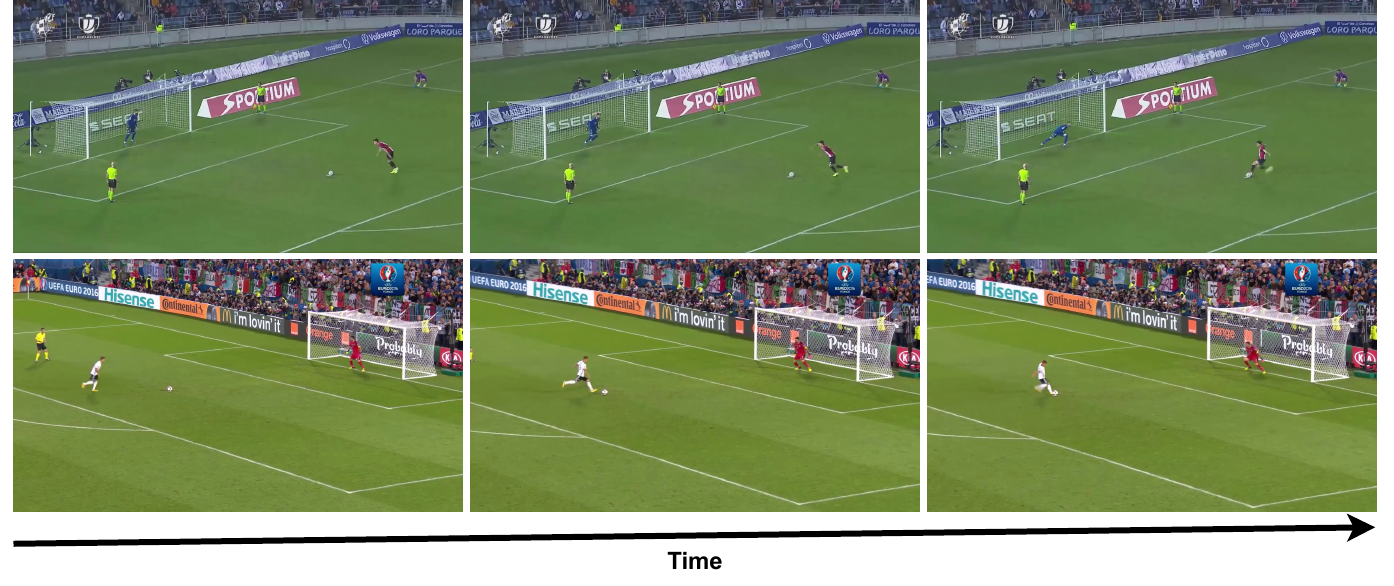}
    \caption{Example frames from the collected penalty-kick dataset, illustrating the diversity in viewpoint, scale, and scene context across clips.}
    \label{fig:datasetimg}
\end{figure}

\section{Experimental Setup}
\label{sec:expSet}

In the experimental setup, the model is trained with a batch size of 5 for up to 60 epochs, using early stopping with a patience of 10 epochs. Optimization is carried out with the AdamW optimizer, using a learning rate of \num{1e-3} and a weight decay of \num{5e-2}. To stabilize training, gradient clipping is applied with a maximum norm of 1.0. The classification objective utilizes cross-entropy loss with class weighting to address class imbalance, and label smoothing is applied with a factor of 0.01. For learning-rate scheduling, a cosine warmup strategy is selected from the considered scheduler options. 

\textbf{Dataset.} As far as it is known, there is no publicly accessible dataset specifically focused on soccer penalty kicks. For this work, the dataset was obtained from material previously compiled in related research~\cite{freire2025predicting}, rather than being collected firsthand. The available recordings come from diverse online sources and display notable variation in body pose, camera distance, angle, and lighting conditions (see Figure~\ref{fig:datasetimg}). As reported by the creators of the dataset, the raw videos were retrieved through targeted web queries related to penalty scenarios, resulting in 1010 initial samples. Each video clip was subsequently edited to retain only the approach of the kicker and the instant of ball impact. The original footage has a resolution of $1280\times720$ and lasts between 3–6 seconds. For this study, each clip is standardized to a 48-frame window: 32 frames capture the run-up, while 8 frames precede and follow the kicking instant. This design omits ball-trajectory information and focuses solely on the kicker’s motion.  Clips with unsuitable camera angles or insufficient length are discarded, yielding 622 valid samples. Player localization is verified using ByteTrack~\cite{zhang2021bytetrack}, and the already mentioned two metadata attributes are included: pitch side and kicking foot. In the final set, 388 kicks originate from the right side of the pitch and 234 from the left, with 486 right-footed and 136 left-footed attempts. Table~\ref{tab:dataset_distrib} summarizes these proportions.

\begin{table}[!h]
    \centering
    \begin{tabular}{ccrrr}
    \toprule
         Metadata & & Left (\%) & Center (\%) & Right (\%) \\
    \midrule
         \multirow{2}{*}{Pitch side} & Right side & 46.65 & 17.78 & 35.57 \\
         & Left side & 48.29 & 14.53 & 37.18 \\
    \midrule     
         \multirow{2}{*}{Kicker foot} & Right-footed & 51.23 & 16.26 & 32.51 \\
         & Left-footed & 33.09 & 17.65 & 49.26 \\
    \bottomrule
    \end{tabular}
    \caption{Dataset percentage distribution based on pitch side and kicker foot metadata.}
    \label{tab:dataset_distrib}
\end{table}

To enhance the generalization ability of the proposed approach, data augmentation is enabled and applied with a probability of 0.90. The augmentation pipeline comprises temporal masking (up to 25\% of the sequence length), temporal shifting (up to 2 frames), and frame dropout (0.08). In addition, additive Gaussian noise (std = 0.012), magnitude jittering (std = 0.04), and feature dropout (0.05) are employed to enhance robustness against measurement noise and feature variability. Finally, metadata noise injection (std = 0.01) is applied to reduce sensitivity to small perturbations in auxiliary inputs.

\textbf{Metric evaluation}. Accuracy (Acc.) is reported together with Precision (P), Recall (R), and F1-score (F1), computed under a 10-fold cross-validation protocol. The F1-score is particularly important in the three-class task due to class imbalance and because the \textit{center} class is typically more ambiguous than \textit{left/right}.

\section{Experimental Results}
\label{sec:expRes}
This section evaluates \textit{MambaKick} for penalty-kick direction estimation under two label spaces: (i) a three-class setting with $\{\textit{left, center, right}\}$ and (ii) a two-class setting with $\{\textit{left, right}\}$ obtained by removing \textit{center} samples. The evaluation aims to measure the benefit of combining pretrained HAR embeddings with an efficient temporal predictor while keeping the visual representation fixed and lightweight.

Tables~\ref{tab:threeclass} and~\ref{tab:twoclass} report performance across a range of HAR architectures. A goalkeeper (GK) baseline is also included and is defined as the direction actually chosen by the goalkeeper during each penalty kick; this provides a behavioral reference point. For each architecture, the table lists the best-performing model variant and the evaluation metrics. Results are reported for (i) reference embedding-based baselines from~\cite{freire2025predicting} and (ii) the proposed \textit{MambaKick} applied to the same HAR embeddings, enabling a fair comparison of temporal modeling.

As shown in Table~\ref{tab:threeclass}, the GK baseline achieves 46.0\% accuracy, highlighting the difficulty of the task under limited reaction time. Among reference embedding baselines, MViTv1 and MViTv2 perform best, reaching 51.9\% and 51.6\% accuracy, respectively. When using \textit{MambaKick}, performance improves or remains competitive across most backbones, with the best accuracy reaching \textbf{53.1\%} (achieved with both MViTv1 and MViTv2 embeddings). These results indicate that modeling the temporal evolution of HAR embeddings can provide consistent gains over static or simpler aggregation strategies.

In Table~\ref{tab:twoclass}, performance increases for all methods due to the reduced label ambiguity. The GK baseline reaches 54.2\% accuracy. Among reference baselines, the strongest result is 63.9\% (MViTv2). The proposed method achieves a best accuracy of \textbf{64.5\%} using MViTv1 embeddings and provides notable improvements for some backbones (e.g., SlowFast), supporting the effectiveness of a lightweight state-space temporal head for exploiting sequential structure in HAR features.

\begin{figure}
\setlength\tabcolsep{1.0pt}
\centering
\scalebox{1.0}{
\begin{tabular}{cc}

\multicolumn{2}{c}{Pitch side} \\
Three classes & Two classes \\
\includegraphics[width=.49\textwidth]{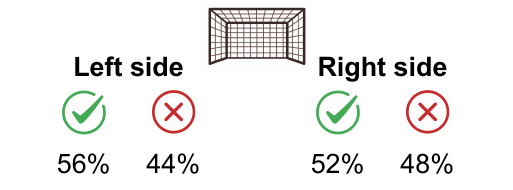} &
\includegraphics[width=.49\textwidth]{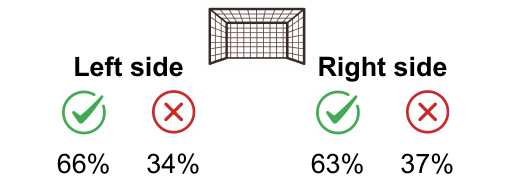} \\

\midrule

\multicolumn{2}{c}{Kicker foot} \\
Three classes & Two classes \\
\includegraphics[width=.49\textwidth]{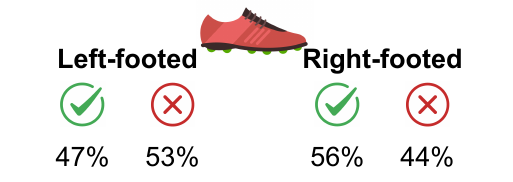} &
\includegraphics[width=.49\textwidth]{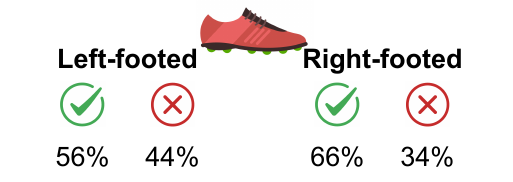} \\
\end{tabular}
}
\caption{Model performance for the best MViTv1 technique (percentage of correct vs. incorrect predictions) stratified by shot placement side (pitch side: left vs. right) and kicker foot (left-footed vs. right-footed), comparing the three-class and two-class classification settings. Percentages indicate accuracy and error rate within each subgroup.}
\label{fig:result_stats}
\end{figure}

Figure~\ref{fig:result_stats} shows analysis using pitch side and kicker foot metadata as a reference for the best MViTv1 technique. The performance improves consistently when moving from the three-class to the two-class setting across all subgroups. For pitch side, accuracy increases from 52\% to 63\% for shots to the right side and from 56\% to 66\% for shots to the left side, indicating a small but consistent advantage for left-side shots in both settings. For kicker foot, the model performs better on right-footed kickers than left-footed kickers: accuracy is 56\% vs. 47\% in the three-class setting and 66\% vs. 56\% in the two-class setting, respectively. Overall, the two-class formulation yields higher accuracy and lower error rates, while the largest performance gap appears between left-footed and right-footed kickers in the three-class task.

\begin{figure}
\setlength\tabcolsep{1.0pt}
\centering
\scalebox{1.0}{
\begin{tabular}{cc}
\includegraphics[width=.49\textwidth]{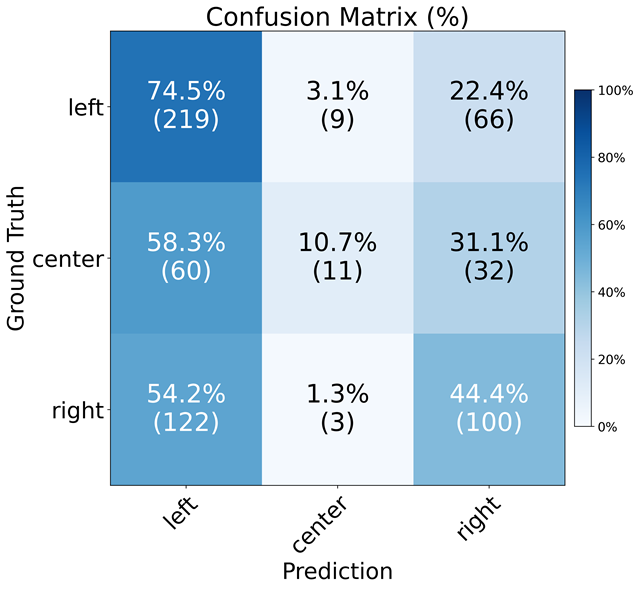} &
\includegraphics[width=.49\textwidth]{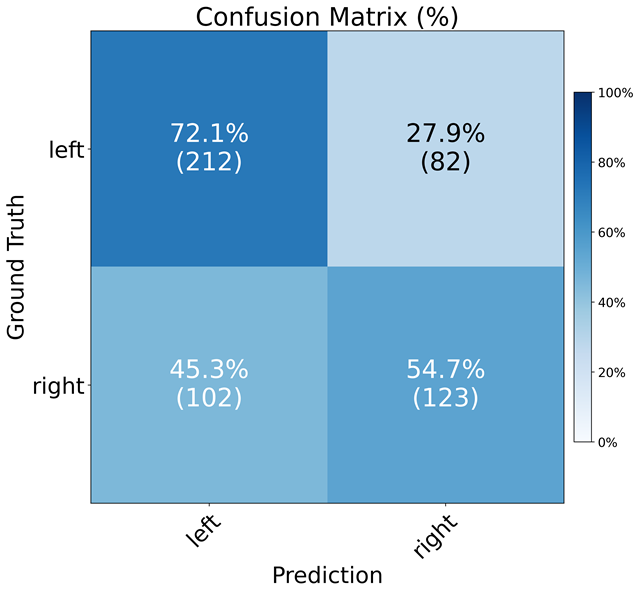} \\
\end{tabular}
}
\caption{(left) Confusion Matrix of 3 classes using MViTv1. (right) Confusion Matrix of 2 classes using MViTv1.}
\label{fig:confmat}
\end{figure}

Finally, Figure~\ref{fig:confmat} shows row-normalized confusion matrices for the best \textit{MambaKick} configuration using MViTv1 embeddings. When using the Three-class setting, the evaluation split contains a total of 622 samples (294 \textit{left}, 103 \textit{center}, and 225 \textit{right}, as reflected in the confusion-matrix counts). The main source of error is the \textit{center} class: only 10.7\% (11/103) of \textit{center} penalties are correctly classified, while most are misclassified as either \textit{left} (58.3\%) or \textit{right} (31.1\%). In contrast, \textit{left} is recognized relatively well (74.5\% recall; 219/294), and \textit{right} achieves moderate recall (44.4\%; 100/225). This behavior suggests that \textit{center} shots are both under-represented and intrinsically difficult to identify from pre-contact cues in unconstrained video, motivating future work on re-balancing strategies and stronger early-prediction objectives. On the other hand, when using the Two-class setting, after removing \textit{center}, the evaluation subset contains 519 samples (294 \textit{left} and 225 \textit{right}). Although overall accuracy improves, errors remain asymmetric: \textit{left} recall is 72.1\% (212/294) compared to 54.7\% (123/225) for \textit{right}. This indicates that ambiguity persists due to factors such as viewpoint variation, occlusions, or mirrored appearances, and may benefit from additional context or viewpoint normalization.

\begin{table}
    \centering
    \begin{tabular}{llcrrrr}
        \toprule
        Architecture & Best Model & Reference & Acc. (\%) & P (\%) & R (\%) & F1 (\%) \\
        \midrule
        GK Baseline & - & - & 46.0 & 35.3 & 44.2 & 38.4 \\
        
        C2D \cite{simonyan2014two} & C2D R50 & Freire-Obreg\'on et al. \cite{freire2025predicting} & 47.1 & 33.1 & 45.6 & 38.4 \\
        I3D \cite{CarreiraZisserman2017QuoVadis} & I3D R50 & Freire-Obreg\'on et al. \cite{freire2025predicting} & 45.0 & 32.7 & 46.3 & 38.3 \\
        Slow \cite{feichtenhofer2021large} & Slow8x8 & Freire-Obreg\'on et al. \cite{freire2025predicting} & 46.7 & 28.9 & 45.9 & 35.5 \\
        SlowFast \cite{Feichtenhofer2019SlowFast} & SlowFast4x16 & Freire-Obreg\'on et al. \cite{freire2025predicting} & 46.2 & 33.6 & \third{46.4} & 39.1 \\
        NLN \cite{wang2018non} & Slow NLN 4x16 & Freire-Obreg\'on et al. \cite{freire2025predicting} & 45.1 & 32.8 & \first{47.8} & 38.9 \\
        X3D \cite{feichtenhofer2020x3d} & X3D M & Freire-Obreg\'on et al. \cite{freire2025predicting} & 45.9 & 33.7 & 45.3 & 38.6 \\
        MViTv1 \cite{fan2021multiscale} & MViT CONV & Freire-Obreg\'on et al. \cite{freire2025predicting} & \second{51.9} & 34.4 & \second{47.1} & 39.7 \\
        MViTv2 \cite{li2022mvitv2} & MViTv2 S & Freire-Obreg\'on et al. \cite{freire2025predicting} & 51.6 & 35.5 & 45.8 & 40.1 \\
        \midrule
        C2D \cite{simonyan2014two} & C2D R50 & MambaKick (Ours) & 48.6 & 34.7 & 37.8 & 35.0 \\
        I3D \cite{CarreiraZisserman2017QuoVadis} & I3D R50 & MambaKick (Ours) & 49.0 & \second{48.8} & 40.8 & \second{41.0} \\
        Slow \cite{feichtenhofer2021large} & Slow8x8 & MambaKick (Ours) & 50.2 & \third{45.8} & 41.3 & 39.8 \\
        SlowFast \cite{Feichtenhofer2019SlowFast} & SlowFast4x16 & MambaKick (Ours) & 51.0 & 44.5 & 39.8 & 37.2 \\
        NLN \cite{wang2018non} & Slow NLN 4x16 & MambaKick (Ours) & \third{51.8} & 42.8 & 41.1 & 39.5 \\
        X3D \cite{feichtenhofer2020x3d} & X3D M & MambaKick (Ours) & 50.5 & 39.5 & 41.6 & 38.7 \\
        MViTv1 \cite{fan2021multiscale} & MViT CONV & MambaKick (Ours) & \first{53.1} & \first{51.0} & 43.2 & \first{42.6} \\
        MViTv2 \cite{li2022mvitv2} & MViTv2 S & MambaKick (Ours) & \first{53.1} & 43.2 & 42.7 & \third{40.8} \\
        \bottomrule
    \end{tabular}
    \caption{
    Experimental results across backbones; a goalkeeper (GK) baseline, defined by the goalkeeper’s chosen direction for three-class, is provided. The best three performing results are highlighted using color: \First{First}, \Second{Second}, and \Third{Third} respectively.}
    \label{tab:threeclass}
\end{table}

\begin{table}
    \centering
    \begin{tabular}{llcrrrr}
        \toprule
        Architecture & Best Model & Reference & Acc. (\%) & P (\%) & R (\%) & F1 (\%) \\
        \midrule
        GK Baseline & - & - & 54.2 & 53.3 & 53.2 & 53.3 \\
        
        C2D \cite{simonyan2014two} & C2D R50 & Freire-Obreg\'on et al. \cite{freire2025predicting} & 58.3 & 58.4 & 55.8 & 57.1 \\
        I3D \cite{CarreiraZisserman2017QuoVadis} & I3D R50 & Freire-Obreg\'on et al. \cite{freire2025predicting} & 55.4 & 55.5 & 54.2 & 54.9 \\
        Slow \cite{feichtenhofer2021large} & Slow8x8 & Freire-Obreg\'on et al. \cite{freire2025predicting} & 61.1 & \first{66.7} & 57.3 & \third{61.6} \\
        SlowFast \cite{Feichtenhofer2019SlowFast} & SlowFast4x16 & Freire-Obreg\'on et al. \cite{freire2025predicting} & 60.5 & 60.4 & 57.7 & 58.7 \\
        NLN \cite{wang2018non} & Slow NLN 4x16 & Freire-Obreg\'on et al. \cite{freire2025predicting} & 61.6 & 60.3 & 58.8 & 59.5 \\
        X3D \cite{feichtenhofer2020x3d} & X3D M & Freire-Obreg\'on et al. \cite{freire2025predicting} & 60.1 & 59.1 & 59.2 & 59.2 \\
        MViTv1 \cite{fan2021multiscale} & MViT CONV & Freire-Obreg\'on et al. \cite{freire2025predicting} & 61.8 & 58.4 & 58.1 & 58.2 \\
        MViTv2 \cite{li2022mvitv2} & MViTv2 S & Freire-Obreg\'on et al. \cite{freire2025predicting} & \third{63.9} & \second{64.9} & 60.2 & \second{62.5} \\
        \midrule
        C2D \cite{simonyan2014two} & C2D R50 & MambaKick (Ours) & 60.1 & 58.9 & 57.7 & 57.3 \\
        I3D \cite{CarreiraZisserman2017QuoVadis} & I3D R50 & MambaKick (Ours) & 60.5 & 59.4 & 59.0 & 59.0 \\
        Slow \cite{feichtenhofer2021large} & Slow8x8 & MambaKick (Ours) & 60.7 & 59.6 & 59.0 & 59.0 \\
        SlowFast \cite{Feichtenhofer2019SlowFast} & SlowFast4x16 & MambaKick (Ours) & \second{64.0} & 63.7 & \third{61.2} & 60.7 \\
        NLN \cite{wang2018non} & Slow NLN 4x16 & MambaKick (Ours) & 61.3 & 60.2 & 59.5 & 59.5 \\
        X3D \cite{feichtenhofer2020x3d} & X3D M & MambaKick (Ours) & 61.3 & 60.2 & 59.2 & 59.1 \\
        MViTv1 \cite{fan2021multiscale} & MViT CONV & MambaKick (Ours) & \first{64.5} & \third{63.8} & \first{63.4} & \first{63.5} \\
        MViTv2 \cite{li2022mvitv2} & MViTv2 S & MambaKick (Ours) & 63.4 & 62.6 & \second{62.5} & \second{62.5} \\
        \bottomrule
    \end{tabular}
    \caption{
    Experimental results across backbones; a goalkeeper (GK) baseline, defined by the goalkeeper’s chosen direction for two-class, is provided. The best three performing results are highlighted using color: \First{First}, \Second{Second}, and \Third{Third} respectively.}
    \label{tab:twoclass}
\end{table}

\textbf{Ablation study.} Table \ref{tab:ablation} shows that in the three-class setting, the model performs worst when using only the Running branch, reaching 49.4\% accuracy and an F1-score of 38.1\%. When the Kicking branch is added (Running + Kicking), performance improves to 52.4\% accuracy and 39.9\% F1-score, suggesting that kicking contributes additional useful information beyond running alone. The best results are achieved when Metadata is included as well (Running + Kicking + Metadata), with 53.1\% accuracy and a notably higher F1-score of 42.6\%, alongside improvements in both precision (51.0\%) and recall (43.2\%). This indicates metadata plays an important role in helping the model separate the more challenging three classes.

Table \ref{tab:ablation} shows that in the two-class setting, results are higher overall. Using only the Running branch already yields strong performance (61.8\% accuracy, 61.1\% F1-score). Similar to two-class, when adding the Kicking branch slightly degrades performance (61.8\% accuracy, 58.2\% F1-score), which may imply that kicking features are more informative for the binary classification task. When Metadata is added (Running + Kicking + Metadata), the model reaches the best performance in this setting, achieving 64.5\% accuracy and 63.5\% F1-score, with corresponding gains in precision (63.8\%) and recall (63.4\%). Overall, running, kicking, and metadata branches provide the most consistent benefit in both scenarios.

\begin{table}
    \centering
    \begin{tabular}{clrrrr}
        \toprule
        \# of class & Branches & Acc. (\%) & P (\%) & R (\%) & F1 (\%) \\
        \midrule

        3 & Running & 49.4 & 40.3 & 39.3 & 38.1 \\
        3 & Running + Kicking & 52.4 & 49.5 & 42.1 & 39.9 \\
        3 & Running + Kicking + Metadata & 53.1 & 51.0 & 43.2 & 42.6 \\
        
        \midrule

        2 & Running & 61.8 & 61.1 & 59.0 & 58.2 \\
        2 & Running + Kicking & 63.0 & 62.1 & 61.3 & 61.4 \\
        2 & Running + Kicking + Metadata & 64.5 & 63.8 & 63.4 & 63.5 \\
        
        \bottomrule
    \end{tabular}
    \caption{Ablation study using MViTv1 \cite{fan2021multiscale} testing branch removal to see the importance of each component.}
    \label{tab:ablation}
\end{table}

\section{Conclusions}
\label{sec:concl}
In conclusion, \textit{MambaKick} is presented as a framework for early penalty-kick direction prediction that leverages pretrained HAR embeddings extracted from a short, contact-centered temporal window and combines them with a lightweight Mamba-based temporal head. Rather than relying on explicit kinematic reconstruction or handcrafted biomechanical cues, the approach reuses transferable spatiotemporal representations and concentrates learning capacity on efficient sequence aggregation, optionally augmented with simple contextual metadata.

Across multiple HAR backbones, \textit{MambaKick} consistently matches or improves strong embedding-based baselines, achieving up to \textbf{53.1\%} accuracy in the three-class (\textit{left/center/right}) setting and \textbf{64.5\%} accuracy in the two-class (\textit{left/right}) setting. Confusion-matrix analysis indicates that the \textit{center} class remains the dominant source of error, with most \textit{center} shots being confused with \textit{left/right}, suggesting that class imbalance and intrinsic visual ambiguity limit performance under real-world conditions.

Overall, the results support the practicality of combining pretrained HAR representations with efficient state-space temporal modeling for low-latency intention prediction in sports video. Future work should address more robust handling of the \textit{center} class (e.g., re-balancing, cost-sensitive losses, or dedicated ambiguity modeling) and evaluate prediction quality as a function of time-to-contact to better characterize true early anticipation performance.

\begin{credits}
\subsubsection{\ackname} This research has been supported by the ESPOL project ``Reconocimiento de patrones en imágenes usando técnicas basadas en aprendizaje'' (CIDIS-004-2024).

\end{credits}
%
%
%
\bibliographystyle{splncs04}
\bibliography{mybibliography}
%




\end{document}